\documentclass[conference]{IEEEtran}
   \usepackage{amsfonts}
   \usepackage[dvipsnames,table,xcdraw]{xcolor}

   \usepackage[pdftex]{graphicx}
  	\graphicspath{{../pdf/}{../jpeg/}}
	\DeclareGraphicsExtensions{.pdf,.jpeg,.png}

    \usepackage[cmex10]{amsmath}
    \usepackage{mathabx}
    \usepackage{algorithm}
    \usepackage{algorithmicx}
    \usepackage[noend]{algpseudocode}

    \usepackage{array}
    \usepackage{mdwtab}
    \usepackage{eqparbox}
    \usepackage{url}
    \usepackage{url}
    \usepackage{float}
    \usepackage{hyperref}
    \usepackage{amsmath}
    \usepackage{multirow}
    \usepackage{booktabs}
    \usepackage{multirow}
    \usepackage{mathtools}
    
    \usepackage{graphicx}
    \usepackage{definitions}

    \usepackage[dvipsnames]{xcolor}
    \usepackage{amssymb,amsthm}
\begin{document}

\title{\LARGE \emph{ElegansNet}: a brief scientific report and initial experiments.} 
 
% \author{\authorblockN{Leave Author List blank for your IMS2013 Summary (initial) submission.\\ IMS2013 will be rigorously enforcing the new double-blind reviewing requirements.}
% \authorblockA{\authorrefmark{1}Leave Affiliation List blank for your Summary (initial) submission}}

 \author{\authorblockN{Francesco Bardozzo\authorrefmark{1}\authorrefmark{3}, Andrea Terlizzi\authorrefmark{1},  Pietro Li\'o \authorrefmark{2}, Roberto Tagliaferri\authorrefmark{1} }
\authorblockA{\authorrefmark{1} \emph{NeuroneLab} - DISA-MIS - University of Salerno - Italy}
\authorblockA{\authorrefmark{2} \emph{Computer Laboratory} - University of Cambridge - United Kingdom}
\authorblockA{\authorrefmark{3}  \scriptsize To whom correspondence should be addressed: \href{}{fbardozzo@unisa.it}}
 
}

\maketitle

\IEEEoverridecommandlockouts
%\begin{keywords}
%C.elegans, neuronal brain, network motif analysis, multi-dyadic effects
%\end{keywords}

\IEEEpeerreviewmaketitle
\begin{abstract}
 This research report introduces ElegansNet, a  neural network that mimics real-world neuronal network circuitry, with the goal of better understanding the interplay between connectome topology and deep learning systems. The proposed approach utilizes the powerful representational capabilities of living beings' neuronal circuitry to design and generate improved deep learning systems with a topology similar to natural networks. The Caenorhabditis elegans connectome is used as a reference due to its completeness, reasonable size, and functional neuron classes annotations. It is demonstrated that the connectome of simple organisms exhibits specific functional relationships between neurons, and once transformed into learnable tensor networks  and integrated into modern architectures, it offers bio-plausible structures that efficiently solve complex tasks. The performance of the models is demonstrated against randomly wired networks and compared to artificial networks ranked on global benchmarks. In the first case, \emph{ElegansNet} outperforms randomly wired networks. Interestingly,  \emph{ElegansNet} models show slightly similar performance with only those based on the Watts-Strogatz small-world property. When compared to state-of-the-art artificial neural networks, such as transformers or attention-based autoencoders, \emph{ElegansNet} outperforms well-known deep learning and traditional models in both supervised image classification tasks and unsupervised hand-written digits reconstruction, achieving  top-1 accuracy of $99.99\%$ on Cifar10 and $99.84\%$ on MNIST Unsup on the validation sets.
\end{abstract}

% ===================
% # I. Introduction #
% ===================

\section{Short Report}
\label{report}
\begin{figure*}[!ht]
\includegraphics[width=1\textwidth]{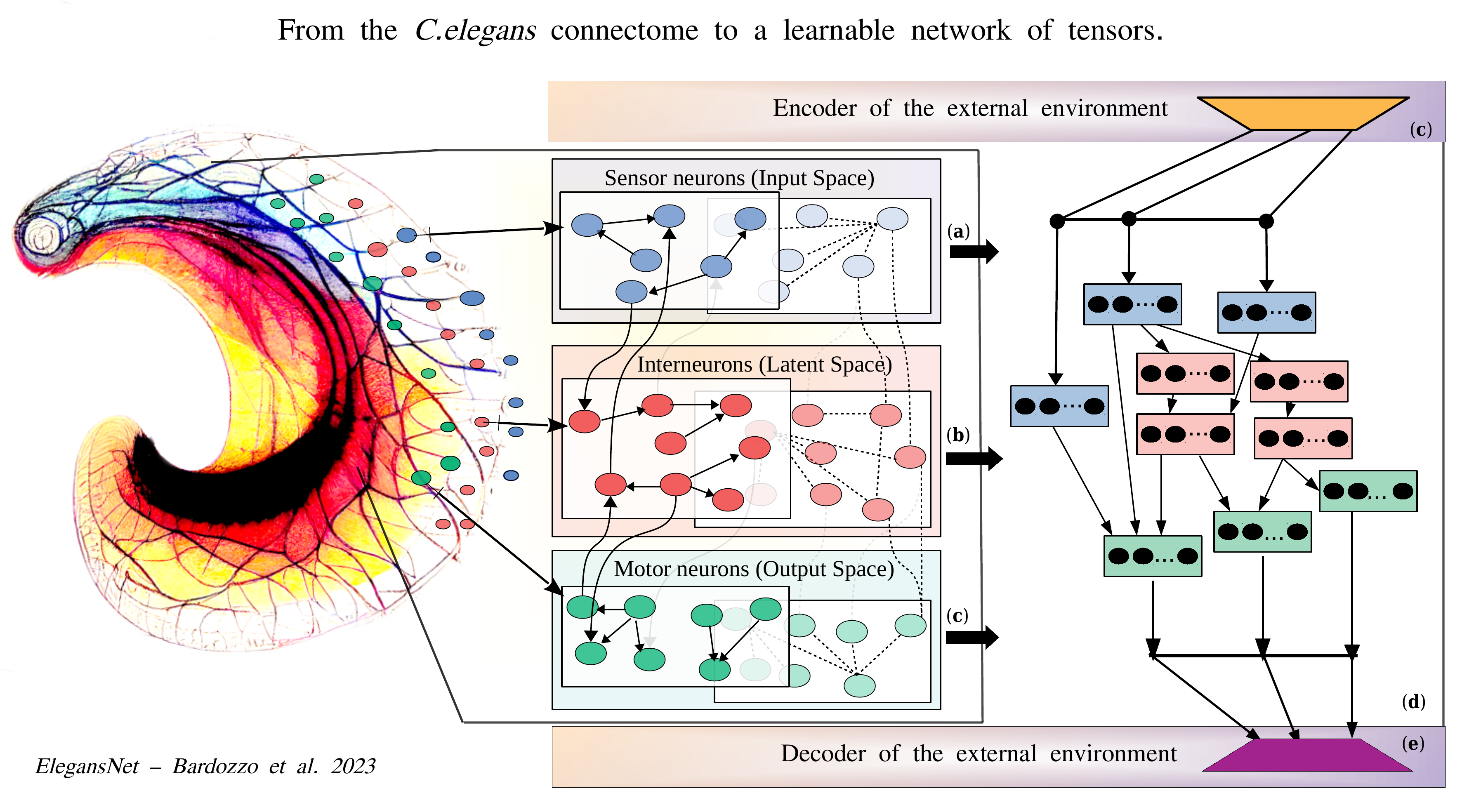}
\caption{ \label{f1}  The connectome of \emph{C.elegans} is represented as a fully connected graph with two overlapping layers, where the solid edges represent chemical and directional synapses and the dashed edges represent electrical and undirected ones. The sensor neurons are represented in blue (Box \textbf{(a)}), while interneurons are represented in red (Box \textbf{(b)}). Finally, the motor neurons are represented in green (Box \textbf{(c)}). The encoder (Box \textbf{(c)}) may differ for various deep learning tasks and is responsible for generating the feature maps that serve as input to the sensor-tensors of the latent space. The learnable network (Box \textbf{(d)}) of tensors is the resulting layered model from the reference graph, projected into the center of the external environment. The tensor network takes the form of a directed acyclic graph and serves as the latent space of our models. In the output of the tensor network, the motor unit tensors are collected, and the most important features are selected in the decoder part (Box \textbf{(d)}).   }
\end{figure*}
For decades, scientists have been working on developing algorithms and machines that are increasingly inspired by the communication mechanisms of the brain and nervous system. In the neural network (connectionist) models, the focus is on artificial network-based architectures that can learn from examples by solving various tasks with adequate generalization capacity \cite{goodfellow2016deep}. Although these modern approaches to problem-solving are widely recognized and applied by the scientific community,  there is still considerable potential for 'bio-inspired' improvements focused on shaping neural networks as connectomes. On the one hand, the brain is a complex network of neurons that work together to perform cognitive tasks, and the study of these networks, known as connectomics, has yielded valuable insights into how the brain functions \cite{fornito2016fundamentals}. Effectively, researchers in connectomics have shown that the topology of neuronal connections is crucial to understanding how the neuronal circuitry processes information \cite{van2017neural}. On the other hand, deep learning has revolutionized the field of artificial intelligence by designing network architectures based on artificial neuronal circuitry, which demonstrates remarkable performance in a wide range of tasks \cite{bardozzo2022stasis, han2021transformer, velickovic2017graph,goodfellow2016deep}. Despite significant progress in their respective fields, researchers working on artificial deep neural networks rarely replicate the topology of the biological neuronal circuitry, also known as the connectome, primarily due to various factors such as limitations in dimensionality and constraints related to cross-domain considerations. Inspired by the work of \cite{xie2019exploring}, a group of researchers proposed the use of Deep Connectomics Networks (DCNs) \cite{roberts2019deep} in 2019. This was the first attempt to design small-world neural networks inspired by real-world neuronal networks in terms of learning capacity. On the counterpart, they did not replicate, in a one-to-one manner, the topology of living being connectomes. As of April 5th, 2023, to the best of our knowledge, the development of DCNs has stalled. To fill this gap, we designed an approach that effectively leverages the strong representational power of a living being's connectome. Our approach is based on designing and generating connectome-inspired learning systems through almost 1:1 transformations of both the neuron representation in tensors and the synapses in tensor connections. Specifically, we use the structural information of the connectome to model the deep learning architecture and the functional organization of sensors, interneurons, and motors to enhance backpropagation-based learning capability \cite{lillicrap2020backpropagation}. Our initial experiments demonstrate that the connectome of the simple organism, \emph{Caenorhabditis elegans} \cite{varshney2011structural}, has specific functional relationships between neurons that have been shaped and optimized by evolutionary pressure, which better form a learnable network of tensors (as briefly discussed in Section \ref{par1} and illustrated in Figure \ref{f1}). By transforming the connectome into a learnable network of tensors and integrating it into modern deep-learning architectures, we have revealed its potential for learning on well-known classification and reconstruction tasks. Figure \ref{f1} provides an example of the transformation process from the connectome to the tensor network. Specifically, the connectome-inspired tensor network consists of a layering procedure that considers sensors as the input space, interneurons as the latent space, and motor neurons as the output space. Thus, the layered connectome is reconstructed from its graph representation using an encoder and decoder, with the inputs and outputs depending on the learning problem. Ideally, the encoder/decoder layers are intended to serve as an external environment where the artificial worm can interact (see Figure \ref{f1} - Box \textbf{(c)} and \textbf{e}). The Sections \ref{subsec:results_random} and \ref{sotacomp} of this report showcase the performance of our \emph{ElegansNet} models in comparison to randomly rewired connectomes and global benchmarks on two well-known problems: \emph{Cifar10} \footnote{\href{https://paperswithcode.com/sota/image-classification-on-cifar-10}{Benchmark Cifar10}}, a supervised problem, and \emph{MNIST-Unsup} \footnote{\href{https://paperswithcode.com/sota/unsupervised-mnist-on-mnist}{Benchmark MNIST Unsup}}\label{mnist_unsup_repo}, an unsupervised problem.

\section{Results}
\subsection{Connectome evolutionary conservation.} 
\label{par1}
The evidence of evolutionary conservation on the reference connectome serves as a crucial optimization key for structuring deep learning models. Therefore, in the first part of our study, we delved deeper into the characteristics of the \emph{C. Elegans} neuronal circuitry, examining how the network's structure affects the relationships between neuronal features. Motif entropy \cite{adami2011information} and dyadic/anti-dyadic algorithms \cite{park2007distribution}, along with their extensions, are used to analyze the distribution of various neuronal functions and their evolution over time. It is noteworthy that Watts and Strogatz \cite{watts1998collective} demonstrated the small-world property of the \emph{C.elegans} connectome, which provides valuable insights into designing networks that are similar to natural ones. This property is also reflected in the performance comparisons discussed in Section \ref{subsec:results_random}. Furthermore, our analysis of the learning capacity of the \emph{C.Elegans} neuronal circuitry suggests that the functional interactions between motor and sensor neurons, as well as interneurons, are preserved characteristics that gradually lose significance as the percentage of rewiring increases. This correlates with the multi-dyadic/anti-dyadic effect, which is a highly-preserved feature as long as the connectomes are not randomized. Even a slight rewiring is enough to disrupt this effect along the shortest paths of length less than 2/3 edges. These results provide valuable insights into structuring graph-optimized deep learning models that emulate the behavior of the connectome.  More generally, understanding how certain topological patterns are distributed throughout the connectome network could provide further clues in the design of deep learning models, as evidenced by some works \cite{Nowak2022:nnrandomgraphs2, janik2020:nnrandomgraphs0, Waqas2021:nnrandomgraphs1}. 

\subsection{Comparisons with randomly generated networks}\label{subsec:results_random}
\begin{figure*}
\centering
       \includegraphics[ height=240px ]{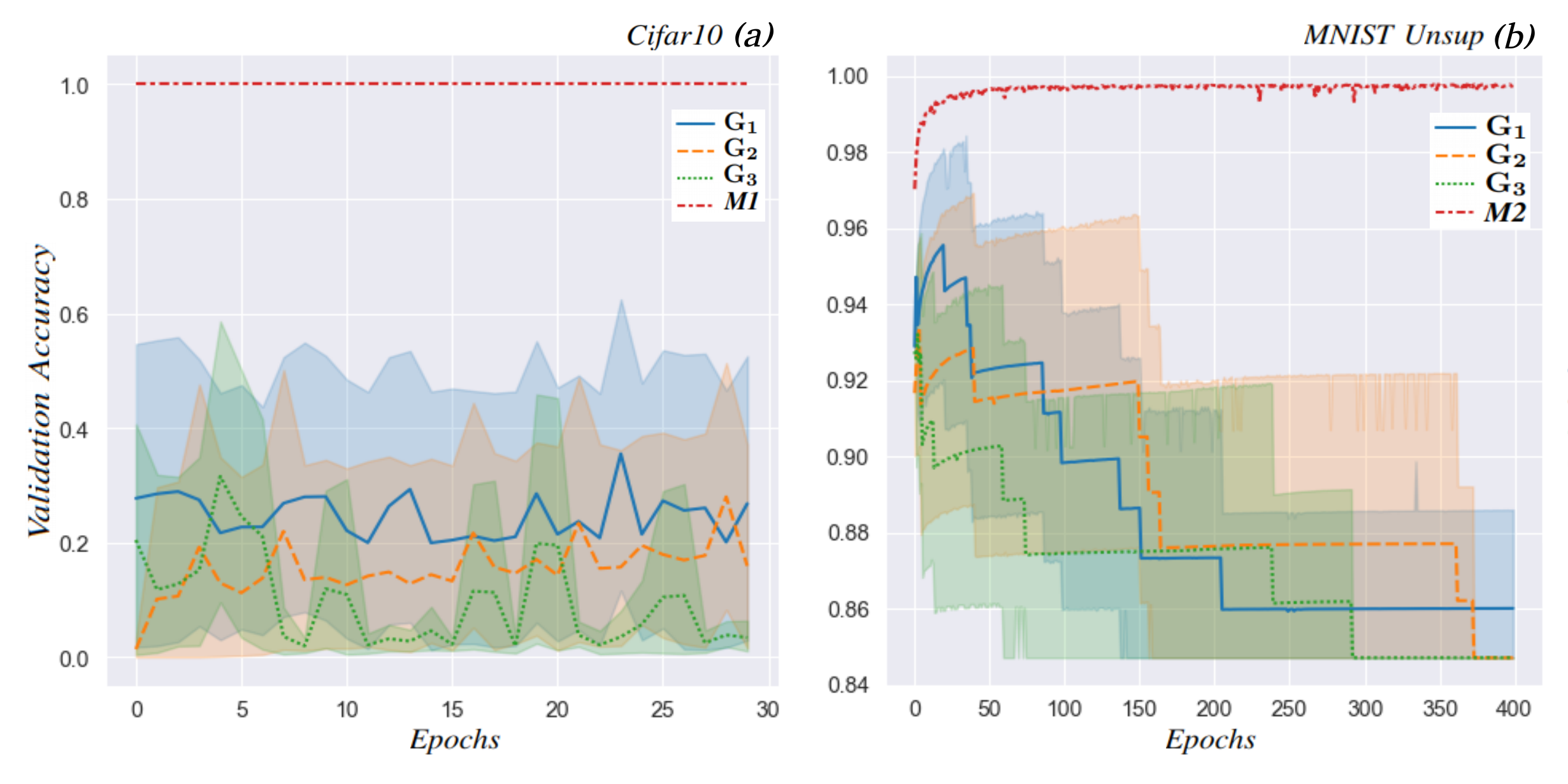}
\caption{\label{acc_cifar} In this Figure, the original neural circuitry forming the \emph{ElegansNet} $M1$ and $M2$ models is compared to randomly rewired models on two well-known problems: \emph{Cifar10} and \emph{MNIST Unsup}, which are shown in boxes \textbf{(a)} and \textbf{(b)}, respectively. The $y$-axis represents the validation set top-1 accuracy, while the $x$-axis represents the number of training epochs.  The models derived from the tensor network of the original connectome are depicted by the red dashed lines, while the models built with randomly rewired small-world graphs (a total of 30) are represented by means of average and variance as follows: $\mathbf{G}_1$ (blue lines) is for Watts-Strogatz, $\mathbf{G}_2$ (orange dotted lines) is for Erdos-Renyi, and $\mathbf{G}_3$ (green dotted lines) is for Barabasi-Albert generators.}
\end{figure*}

This section compares randomly wired small-world networks with the natural network organization of the \emph{C.Elegans} connectome from a deep learning modeling perspective. Unlike other works \cite{xie2019exploring, roberts2019deep, janik2020:nnrandomgraphs0, Waqas2021:nnrandomgraphs1, Nowak2022:nnrandomgraphs2},  an algorithm that directly converts the input graph nodes into neural network tensors is employed. Such tensors are functionally related to each other according to the edges of the aforementioned graph. As a result, no starting architecture (e.g. ResNet for \cite{xie2019exploring, roberts2019deep} is required, and an almost 1:1 mapping between the tensors and connections of the generated network and the source graph is produced. The model thus constructed, based on the \emph{C.elegans} natural connectome, is referred to as \textit{ElegansNet}. In the first comparison on \emph{Cifar10} classification task, \emph{ElegansNet} $M1$ quickly reaches the $99.99\%$ of accuracy with respect to the randomly generated ones (see also Figure \ref{acc_cifar} - Box \textbf{(a)}), remaining consistent across epochs, outperforming all 30 models based on randomized small-world graphs. As shown in Figure \ref{acc_cifar} - Box \textbf{(a-b)}, the models structured with the Watts-Strogatz (WS) \cite{watts1998collective} generative algorithm exhibit slightly better performance, on average, compared to those structured with the Barabasi-Albert (BA) \cite{albert2002statistical} (green dotted lines) or Erdos-Renyi (ER) \cite{erdHos1960evolution} (orange dotted lines) generative algorithms in both \emph{Cifar10} (ElegansNet $M1$)  and \emph{MNIST}-Unsup (ElegansNet $M2$) training. Conversely, models structured with the ER algorithm in Figure \ref{acc_cifar} - Box \textbf{(b)} maintain a higher accuracy in subsequent epochs. 
These findings further emphasize the significance of natural network connections, highlighting their differences from random ones, and providing insights for the design of more effective artificial neural networks.

\subsection{Comparisons with state-of-the-art models}\label{sotacomp} 
As shown in Table \ref{t1},  \emph{ElegansNet} \emph{M1} reached the $99.99\%$ of top-1 accuracy on the \emph{Cifar10} classification task. Thus, \emph{ElegansNet} \emph{M1} outperforms various SOTA models which include classical vision transformer architectures like \emph{ViT, CvT, CaiT, BiT} or \emph{DeiT} \cite{vaswani2017attention, dosovitskiy2020image, touvron2021going, kolesnikov2020big, touvron2021training, wu2021cvt}, evolutionary-based transformer approaches like $\mu$2Net \cite{gesmundo2022evolutionary}, and pure convolutional architectures like \emph{EfficientNetV2} \cite{tan2021efficientnetv2}. It is worth noting that despite having much fewer trainable parameters ($107M$) than the best transformer, \emph{ViT-H 14} \cite{dosovitskiy2020image}, which has \emph{623M} parameters, \emph{ElegansNet} $M1$ still outperformed it, as well as \emph{EfficientNetV2-L} \cite{tan2021efficientnetv2}, with $121M$ parameters and the \emph{ResNet}-inspired transformer \emph{BiT-L} \cite{kolesnikov2020big}.

\begin{table}[!ht]
\centering
\caption{\label{t1} \emph{ElegansNet} $M1$ vs SOTA models for \emph{Cifar10}}
% Please add the following required packages to your document preamble:
% \usepackage[table,xcdraw]{xcolor}
% If you use beamer only pass "xcolor=table" option, i.e. \documentclass[xcolor=table]{beamer}
\begin{tabular}{|l|c|c|}
\hline
\rowcolor[HTML]{EFEFEF} 
\textbf{Model} &
  \multicolumn{1}{l|}{\cellcolor[HTML]{EFEFEF}\textbf{ top-1Acc.}} &
  \multicolumn{1}{l|}{\cellcolor[HTML]{EFEFEF}\textbf{Param.}} \\ \hline
\textbf{ElegansNet} $M1$ \textbf{(ours)}                                            & \textbf{99.9}                 & \multicolumn{1}{c|}{107M} \\ \hline
\cellcolor[HTML]{FFFFFF}{\color[HTML]{212529} ViT-H/14 \cite{dosovitskiy2020image}} &
  99.5 &
  \cellcolor[HTML]{FFFFFF}{\color[HTML]{212529} 632M} \\ \hline
\rowcolor[HTML]{FFFFFF} 
{\color[HTML]{212529} $\mu$2Net \cite{gesmundo2022evolutionary}}   & {\color[HTML]{212529} 99.5}  & {\color[HTML]{212529} }        \\ \hline
\rowcolor[HTML]{FFFFFF} 
{\color[HTML]{212529} ViT-L/16 \cite{dosovitskiy2020image}}    & {\color[HTML]{212529} 99.4}  & {\color[HTML]{212529} 307M}    \\ \hline
\rowcolor[HTML]{FFFFFF} 
{\color[HTML]{212529} CaiT-M-36 U 224 \cite{touvron2021going}} & {\color[HTML]{212529} 99.4}   & {\color[HTML]{212529} }        \\ \hline
\rowcolor[HTML]{FFFFFF} 
{\color[HTML]{212529} CvT-W24 \cite{wu2021cvt}}                & {\color[HTML]{212529} 99.4}  & {\color[HTML]{212529} }        \\ \hline
\rowcolor[HTML]{FFFFFF} 
{\color[HTML]{212529} BiT-L \cite{kolesnikov2020big}}          & {\color[HTML]{212529} 99.4}  & {\color[HTML]{212529} }        \\ \hline
\rowcolor[HTML]{FFFFFF} 
{\color[HTML]{212529} ViT-B \cite{touvron2022three}}           & {\color[HTML]{212529} 99.3}   & {\color[HTML]{212529} }        \\ \hline
\rowcolor[HTML]{FFFFFF} 
{\color[HTML]{212529} Heinsen Routing + BEiT-large 16 224 \cite{heinsen2022algorithm}} &
  {\color[HTML]{212529} 99.2} &
  {\color[HTML]{212529} 309.5M} \\ \hline
\rowcolor[HTML]{FFFFFF} 
{\color[HTML]{212529} ViT-B/16 \cite{tseng2022perturbed}}      & {\color[HTML]{212529} 99.1} & {\color[HTML]{212529} }        \\ \hline
\rowcolor[HTML]{FFFFFF} 
{\color[HTML]{212529} CeiT-S \cite{yuan2021incorporating}}     & {\color[HTML]{212529} 99.1}   & {\color[HTML]{212529} }        \\ \hline
\rowcolor[HTML]{FFFFFF} 
{\color[HTML]{212529} AutoFormer-S 384 \cite{chen2021autoformer}} &
  {\color[HTML]{212529} 99.1} &
  {\color[HTML]{212529} 23M} \\ \hline
\rowcolor[HTML]{FFFFFF} 
{\color[HTML]{212529} TNT-B \cite{han2021transformer}}         & {\color[HTML]{212529} 99.1}   & {\color[HTML]{212529} 65.6M}   \\ \hline
\rowcolor[HTML]{FFFFFF} 
{\color[HTML]{212529} DeiT-B \cite{touvron2021training}}       & {\color[HTML]{212529} 99.1}   & {\color[HTML]{212529} 86M}     \\ \hline
\rowcolor[HTML]{FFFFFF} 
{\color[HTML]{212529} EfficientNetV2-L \cite{tan2021efficientnetv2}} &
  {\color[HTML]{212529} 99.1} &
  {\color[HTML]{212529} 121M} \\ \hline
\end{tabular}
\end{table}

\emph{ElegansNet} $M2$ is configured to reconstruct handwritten digits (MNIST Unsup). Also in this case the model outperforms machine/deep learning-based SOTA models in global benchmarks. These include a wide range of machine learning techniques, varying from autoencoder-like architectures like Stacked Capsule Autoencoders or Adversarial Autoencoders \cite{kosiorek2019stacked, makhzani2015adversarial}, to GAN-based methods like CatGAN, InfoGAN or PixelGAN \cite{springenberg2015unsupervised, hinz2018inferencing, makhzani2017pixelgan}, to information theory and topology-based algorithms, like Invariant Information Clustering (IIC) and Sparse Manifold Transform \cite{ji2019invariant, chen2022minimalistic}. Table \ref{t2} shows our \emph{ElegansNet} $M2$ results in comparison with both deep and traditional machine learning problems. As indicated in the Section \ref{report}, all the showed results are collected from the online benchmark repository except for Stacked Capsule AutoEncoder (AE) \cite{kosiorek2019stacked}, where instead of performance on MNIST, an accuracy of $98.2$ on the $40\times40$ \textit{MNIST} dataset is reported. \emph{ElegansNet} reaches a \emph{ top-1 accuracy} value of $99.78$ with \emph{MSE} equal to $0.0018$, overreaching all the other models in the chart for which the metric is reported. For what is concerning \emph{F1 score} is $99.27$ overreaching \emph{DenMune} \cite{abbas2021denmune} model that shows a score of $96.6$. The proposed model shows very promising results, improving unsupervised digit reconstruction performances and offering a new approach to designing and understanding deep learning architectures.

% Please add the following required packages to your document preamble:
% \usepackage[table,xcdraw]{xcolor}
% If you use beamer only pass "xcolor=table" option, i.e. \documentclass[xcolor=table]{beamer}
\begin{table}[!ht]
\centering
\caption{\label{t2} \emph{ElegansNet} $M2$ vs SOTA models for MNIST Unsup}
\begin{tabular}{|l|c|}
\hline
\multicolumn{1}{|c|}{ \textbf{Model}} & \textbf{ top-1 Acc.}  \\ \hline
\textbf{ElegansNet $M2$ (ours)} & \textbf{99.8}      \\ \hline
IIC  \cite{ji2019invariant}             & 99.3 \\ \hline
{Sparse Manifold Transform} \cite{chen2022minimalistic}  & {99.3} \\ \hline

{SubTab \cite{ucar2021subtab}}                     & {98.3}         \\ \hline

Stacked Capsule Autoencoder \cite{kosiorek2019stacked}      & {98.0}         \\ \hline

{Self-Organizing Map \cite{khacef2020improving}}   & {96.9}          \\ \hline

{Bidirectional InfoGAN \cite{hinz2018inferencing}} & {96.6}         \\ \hline

{Adversarial Autoencoder \cite{makhzani2015adversarial}}    & {95.9}          \\ \hline

{CatGAN \cite{springenberg2015unsupervised}}       & {95.7}         \\ \hline

{InfoGAN \cite{chen2016infogan}}                   & {95.0}            \\ \hline

{PixelGAN AE \cite{makhzani2017pixelgan}}          & {94.7}         \\ \hline

\multicolumn{1}{|c|}{ \textbf{Model}} & \textbf{F1 (\%)}  \\ \hline
\textbf{ElegansNet $M2$ (ours)}                   & \textbf{99.3}      \\ \hline
DenMune  \cite{abbas2021denmune}             & 96.6 \\ \hline
\end{tabular}

\end{table}

 \subsection{Conclusions}
 \label{subsec:results_conclusions}
 As shown in this preliminary research report \footnote{As this is a preliminary report, there are no details on the evolutionary connectome optimization and how it is structured to represent a connectome neural network.}, these findings suggest that designing models based on the natural graph structure of the \emph{C.Elegans} connectome can be a promising approach for solving complex tasks.  Two comparisons between connectome-inspired and artificial neural networks reveal the remarkable complexity, efficiency, robustness, and flexibility of the former.  Further research is necessary to fully understand neuronal circuitry functions and to develop more sophisticated artificial networks that can better approximate the complexity and adaptability of their biological counterparts. On one hand, despite significant advances in artificial networks, they still fall short of matching the elevated capabilities of biological networks. On the other hand, the integration of biological structures in artificial neural networks is becoming increasingly feasible due to novel approaches such as our \emph{ElegansNet} learning system. This work opens up new avenues of research that were previously out of reach, and the future of neural network research is anticipated to involve greater integration of biological and artificial systems, leading to novel insights and breakthroughs in the field.

%Magnitudine diadica e antidiafica per il modello di park e barabasi (a), MIDEA a shortest path 4 non diretto, MIDEA a shortest path 3 diretto. \emph{Legend}:
% \emph{m} = dyad/multi-dyad effect 
% \emph{s} = sensory neuron (in \textcolor{RoyalBlue}{blue}) 
% \emph{m} = motor neuron   (in \textcolor{WildStrawberry}{red}) 
% \emph{i} = interneuron    (in \textcolor{ForestGreen}{green}) Reference Connectome - \emph{RC} (\textcolor{red}{Red} dot)   
%Violin plots of 5000 rewired connectomes  with edge swap probability of : 
% 1) 0.2 (\textcolor{BurntOrange}{Orange} colour)  
% 2) 0.4  (\textcolor{Goldenrod}{Yellow} colour)  
% 3) 0.6 (\textcolor{PineGreen}{PineGreen} colour) 
% 4) 0.8  (\textcolor{BrickRed}{BrickRed} colour)    
% 5) 1.0 (\textcolor{Rhodamine}{Rhodamine}  colour) 

% ==================
% # ACKNOLEDGMENTS #
% ==================
%We thank in advance San Filippo Neri.
% use section* for acknowledgement
%\section*{Acknowledgment}
% The authors would like to thank...

% ==============
% # REFERENCES #
% ==============

\bibliographystyle{IEEEtran}
\bibliography{IEEEabrv,biblio}

\end{document}